\pdfoutput=1

\documentclass[11pt]{article}

\usepackage[preprint]{acl}

\usepackage{times}
\usepackage{latexsym}

\usepackage[T1]{fontenc}

\usepackage[utf8]{inputenc}

\usepackage{microtype}

\usepackage{inconsolata}

\usepackage{graphicx}
\usepackage{microtype}
\usepackage{subfigure}
\usepackage{booktabs} 
\usepackage{multirow}
\usepackage{hyperref}

\usepackage{makecell}

\usepackage{amsmath}
\usepackage{amssymb}
\usepackage{mathtools}
\usepackage{amsthm}
\usepackage{physics}
\usepackage{array}
\usepackage{xcolor}
\usepackage{adjustbox}
\usepackage{comment}

\DeclareMathOperator*{\argmin}{argmin}

\newcommand\blfootnote[1]{%
\begingroup 
\renewcommand\thefootnote{}\footnote{#1}%
\addtocounter{footnote}{-1}%
\endgroup 
}

\title{Consecutive Batch Model Editing with HooK Layers \thanks{\hspace{0.3mm} This research was substantially supported by the Center for Perceptual and Interactive Intelligence (CPII) Ltd. under the Innovation and Technology Commission’s InnoHK scheme.}}

\author{First Author \\
  Affiliation / Address line 1 \\
  Affiliation / Address line 2 \\
  Affiliation / Address line 3 \\
  \texttt{email@domain} \\\And
  Second Author \\
  Affiliation / Address line 1 \\
  Affiliation / Address line 2 \\
  Affiliation / Address line 3 \\
  \texttt{email@domain} \\}

\author{Shuaiyi Li$^1$, Yang Deng$^{2,\dagger}$, Deng Cai$^{3}$, Hongyuan Lu$^{1}$, Liang Chen$^{1}$, Wai Lam$^1$ \\
  $^1$ The Chinese University of Hong Kong, $^2$ Singapore Management University, $^3$ Tencent AI Lab\\
  \texttt{sli@se.cuhk.edu.hk},  \texttt{ydeng@smu.edu.sg},  \texttt{thisisjcykcd@gmail.com}, \\
  \texttt{1155167604@link.cuhk.edu.hk},  \texttt{lchen@se.cuhk.edu.hk},  \texttt{wlam@se.cuhk.edu.hk}}

\begin{document}
\maketitle
\begin{abstract}
As the typical retraining paradigm is unacceptably time- and resource-consuming, researchers are turning to model editing to find an effective way that supports both consecutive and batch scenarios to edit the model behavior directly. Despite all these practical expectations, existing model editing methods fail to realize all of them. Furthermore, the memory demands for such sequential model editing approaches tend to be prohibitive, frequently necessitating an external memory that grows incrementally over time. To cope with these challenges, we propose CoachHooK, a model editing method that simultaneously supports sequential and batch editing. CoachHooK is memory-friendly as it only needs a small amount of it to store several hook layers whose size remains unchanged over time. Experimental results demonstrate the superiority of our method over other batch-supportive model editing methods under both single-round and consecutive batch editing scenarios. Extensive analyses of CoachHooK have been conducted to verify the stability of our method over a number of consecutive steps.
\blfootnote{$^\dagger$ Corresponding author.}
\end{abstract}

\section{Introduction}
\label{sec.Introduction}
Large Language Models (LLMs) \cite{flan-t5,gpt-4,gpt-neox-20b,llama2} have been demonstrated to be capable of recalling factual knowledge about the real world \cite{brown, Petroni}. Nevertheless, researchers also reveal that LLMs often fail to recall the most up-to-date knowledge or information and some specialized knowledge if they are not periodically updated \cite{AdSZYG22, AgarwalN22, Lazaridou}. Despite the fact that fresh and customizable knowledge is highly desired in many areas, such as text generation, question-answering, reasoning, etc., updating the model via retraining is both time- and resource-consuming. Additionally, researchers have uncovered that well-trained LLMs do make mistakes. One popular sort of mistake is called hallucination \cite{abs-2401-01313}, which means that LLMs generate text based on "hallucinated" fake knowledge. Although many researchers have tried to mitigate this issue \cite{QiuZKPC23, abs-2305-15852, abs-2311-09114, abs-2307-03987}, the strategy to fix this bug remains unclear. Therefore, researchers have started to seek an efficient approach that could edit LLMs in a customizable, cost-effective way.

To this end, recent years have witnessed many efforts in investigating the model-editing techniques to bypass the retraining paradigm and edit the LLMs directly \cite{rome, grace, pmet, mend, serac}. Accordingly, several new datasets (\textit{e.g.}, ZsRE \cite{zsre} and COUNTERFACT \cite{memit}) and corresponding metrics (\textit{e.g.}, reliability, generality, locality, portability \cite{Yao}) are proposed to facilitate the development in this field.
However, these methods either require extra expensive training of a meta-network \cite{mend, ke}, or a classifier \cite{serac}, which causes time and resources overhead, or demands an external memory of explicit edit instances for reference \cite{serac, grace}, which inevitably escalates the memory requirement. Further, most of existing methods are only 
evaluated on single-round editing, where the model is rolled back to the initial state after each edit step.
This deviates from the application scenario in reality since most users anticipate an editing approach that 
allows sequential and batch editing.

In light of these issues, we propose a novel method, named \textbf{CoachHooK}\footnote{Code is released at \url{https://github.com/Syon-Li/CoachHook}.}, which performs \textbf{Co}nsecutive B\textbf{a}t\textbf{ch} Model Editing with \textbf{HooK} layers.
Specifically, CoachHooK supports consecutive batch editing and utilizes the hook layers to separate weight change from the original model weight. 
CoachHook does not need any training on parameters or large explicit external memory that stores editing instances. It only needs a reasonable amount of memory to collect the optimized weight in the hook layer. 
To achieve this, we propose a new transformer memory updating mechanism that supports consecutive batch editing settings and design a simple yet effective local editing scope identification technique used in the hook layer that can accurately detect the inputs in the local editing scope.
We demonstrate the effectiveness of our method via extensive experiments on ZsRE and COUNTERFACT datasets using two popular autoregressive language models, GPT2-XL and GPT-J (6B). Both the single-round batch settings and consecutive batch settings are included, with the total number of editing instances ranging from 1k to 10k. An analysis of the editing scope identification has also been conducted to validate the method. Beyond all these, we implement comprehensive ablation studies to verify the validity of each component and discuss the optimal hyperparameter settings in the method.

\section{Preliminaries of Model Editing}

As defined by \citet{Yao}, the task of model editing is to efficiently modify an initial base model $f_{\theta}$ into an edited model $f_{\theta'}$ whose responses to a particular set of input instances $\mathcal{X}_t$ are adjusted as desired without affecting the responses of the model to other instances. The intended edit descriptor is denoted as $(x_t,y_t)$, where $x_t\in\mathcal{X}_t$ and $f_{\theta}(x_t)\neq y_t$. The post-edit model $f_{\theta'}$ is supposed to produce the expected output to an intended edit instance $x_t$, while preserving the original output to other instances:
\begin{equation}
    f_{\theta'}(x) = \left\{\begin{array}{l}
    y_t   \quad \quad ~~ \text{if} ~ x\in\mathcal{X}_t\\
    f_{\theta}(x) \quad \text{if} ~ x\notin \mathcal{X}_t
    \end{array}\right. 
\end{equation}

In particular, there are three standard criteria for model editing, namely Reliability, Generality, and Locality \cite{Yao, mend, serac}. 
Suppose the prediction of the original model to the prompt "\textit{What is the native language of Joe Biden?}" is "\textit{French}", and 
the expected post-edit model prediction is "\textit{English}".
To verify the Reliability, we use the same original prompt as input and then assess whether the post-edit model predicts "\textit{English}" as desired.
For Generality, a rephrased prompt "\textit{The mother tongue of Joe Biden is"} could be inputted into the edited model to assess whether the output of the model remains as "\textit{English}". Locality 
suggests that the model output of an irrelevant prompt like "\textit{What is the native language of Donald Trump?}" should remain unaffected, which means that the post-edit model should output whatever the initial model output to this prompt. 

The current problem settings of model editing can be generally categorized into three groups \cite{Yao}: 

\noindent1) \textbf{Single instance Editing} evaluates the post-edit model performance when only one single knowledge update is performed:
\begin{equation}
    \theta' \leftarrow \argmin\nolimits_\theta (\parallel{f_{\theta}(x_t)-y_t}\parallel)
\end{equation}

\noindent2) \textbf{Batch Editing} evaluates the post-edit model performance in a more realistic scenario where multiple knowledge pieces are modified simultaneously:
\begin{equation}
    \theta' \leftarrow \argmin\nolimits_\theta \sum\nolimits_{t=1}^{n}(\parallel{f_{\theta}(x_t)-y_t}\parallel)
\end{equation}
where $n\leq \mid{\mathcal{X}_t}\mid$ is the batch size and it varies for different methods \cite{memit,mend,serac,rome}. 

\noindent3) \textbf{Sequential Editing} requires every single edit to be performed successively, and evaluation has to be conducted after a series of knowledge updates \cite{grace}:
\begin{equation}
{\theta_{\mid{\mathcal{X}_t}\mid}}' \leftarrow \argmin\nolimits_{\theta_t} \sum\nolimits_{t=1}^{\mid{\mathcal{X}_t}\mid}(\parallel{f_{\theta_t}(x_t)-y_t}\parallel) 
\end{equation}

In this work, we investigate a new and more practical setting for model editing, namely \textbf{Consecutive Batch Editing}, which aims at executing the editing in a consecutive batch editing way:
\begin{equation}
    \begin{split}
        {\theta_{\lceil \mid{\mathcal{X}_t}\mid/n \rceil}}' &\leftarrow \argmin_{\theta_s} \sum_{s=0}^{\lceil \mid{\mathcal{X}_t}\mid/n \rceil} \Big[\\ &\sum_{t=s \cdot n}^{\min((s+1) \cdot n, \mid{\mathcal{X}_t}\mid)}(\parallel{f_{\theta_s}(x_t)-y_t}\parallel) \Big]
    \end{split}
\end{equation}
where $s$ represents the consecutive editing step.

\section{Method}


We first discuss our method under the single-layer consecutive batch editing setting. Explicitly, we first discuss the process of extending the single-layer updating mechanism in MEMIT \cite{memit} from a scenario of single-round batch editing to consecutive batch editing. Then, we 
introduce the hook layer and the local editing scope identification operation employed in the hook layer. The practicality of the operation is also clarified. Finally, we broaden the method from single-layer to multi-layer scenarios.

\subsection{Single-Layer Consecutive Batch Editing}



\subsubsection{Batch Editing Menchanism}
\citet{memit} demonstrate an effective single-layer editing method using minimal squared error. Although it supports multiple edits on a single round, the updates do not account for scenarios involving consecutive updates. In this section, we extend this approach to include consecutive scenarios. 
Following \cite{memit,rome}, we analyse the model layer weights $W_0$ as a linear associative memory \cite{Kohonen72,anderson1972simple} that stores associations between a set of keys $k_i$ and values $v_i$ using minimal squared error:
\begin{gather}
W_0 = \argmin\nolimits_{W} \sum\nolimits_{i=1}^n\norm{Wk_i - v_i}^2 \label{eq.1}
\end{gather}
In this work, $W_{0}$ is the weight of the second layer of the model's FFN part (denoted as $W_{proj}^l$). For simplicity, we stack keys and values into matrices $K_0=[k_1|k_2|...|k_n]$ and $V_0=[v_1|v_2|...|v_n]$, then Eq. \ref{eq.1} can be optimized by solving \cite{algebra}:
\begin{gather}
W_0K_0K_0^T = V_0K_0^T \label{eq.2} 
\end{gather}
Thanks to the well-conducted pre-training procedure for most of the available LLMs, we can assume that the pre-trained weight $W_{0}$ satisfies Eq. \ref{eq.2}, \textit{i.e.}, serves as the optimal solution for Eq. \ref{eq.1}. 

Unlike \citet{memit}, we define a successive mass-editing objective:
\begin{gather}
\begin{split}
\hat{W_1} = \argmin\nolimits_{W} (&\sum\nolimits_{i=1}^r\norm{Wk_i - v_i}^2 \\
&+ \sum\nolimits_{i=r+1}^{r+u}\norm{Wk_i - v_i}^2) \label{eq.3} 
\end{split}
\end{gather}
Following Eq. \ref{eq.2}, we conclude that Eq. \ref{eq.3} can be optimized if we can solve:
\begin{gather}
\hat{W_1}[K_1\, K_2][K_1\, K_2\,]^T\, =\, [V_1\, V_2][K_1\, K_2]^T \label{eq.4} 
\end{gather}
where $K_1=[k_{1}|k_{2}|\dots|k_{r}] (r \geq n)$ and $V_1=[v_{1}|v_{2}|\dots|v_{r}]$ is the set of key-value pairs that have been updated and $K_2=[k_{r+1}|k_{r+2}|...|k_{r+u}]$ and $V_2=[v_{r+1}|v_{r+2}|...|v_{r+u}]$ is the set of key-values that are going to be edited. Therefore, the objective (Eq. \ref{eq.3}) indicates that we want an optimal $\hat{W_1}$ that successfully updates the new associations while maintaining the old key-value pairs.

Further expanding Eq. \ref{eq.4}:
\begin{gather}
(W_1+\Delta)(K_1K_1^T+K_2K_2^T)=(V_1K_1^T+V_2K_2^T) \\
\begin{split}
W_1K_1K_1^T &+ \Delta K_1K_1^T + W_1K_2K_2^T\\
&+\Delta K_2K_2^T=V_1K_1^T+V_2K_2^T 
\end{split}
\label{eq.6}
\end{gather}
The $\Delta$ means the desired weight change to update the new associations $K_2,V_2$ and $W_1$ is the weight that has been updated for the associations $K_1,V_1$ (Note that $W_1=W_0$ if and only if $r=n$). In a real consecutive editing scenario, $r$ increases and starts with $n$, and each batch-editing iteration is optimized through the objective (Eq. \ref{eq.3}). Hence, we can conclude that $W_1K_1K_1^T=V_1K_1^T$.
Subtracting it from Eq. \ref{eq.6}, we get:
\begin{equation}
\Delta K_1K_1^T + W_1K_2K_2^T + \Delta K_2K_2^T=V_2K_2^T
\end{equation}
Further rearranging it, we have:
\begin{gather}
\Delta = RK_2^TC_{accu}^{-1} \label{eq.8} 
\end{gather}
where $R=(V_2-W_1K_2)$ is the residual error evaluated on the most recent updated weights. $C_{accu}=(K_1K_1^T+K_2K_2^T)$ is the accumulation sum of editing keys' outer product, and we have
\begin{gather}
K_1K_1^T = K_0K_0^T + K'K'^T 
\label{eq.14}
\end{gather}
where $K_0$ is the set of pre-training keys that have been contained in the pre-training weight, $K'=[k_{n+1}|k_{n+2}|...|k_r]$ denotes the updated keys proceeding to current editing step. We follow \cite{memit} to model $K_0K_0^T$ as the uncentered covariance of some randomly sampled inputs:
\begin{gather}
K_0K_0^T = \lambda E[kk^T] 
\end{gather}
Note that the $\lambda$ represents a factor that balances the pre-trained and the whole updated associations.
We follow the definitions of keys and values in \cite{memit, rome}, where keys are the activations at the last token of the subject (such as "Joe Biden" for example provided in \S \ref{sec.Introduction}) and values are gradient-descent optimized vectors that maximize the model's prediction for the target object.

\begin{figure}[t]
\centering
\includegraphics[width=0.48\textwidth]{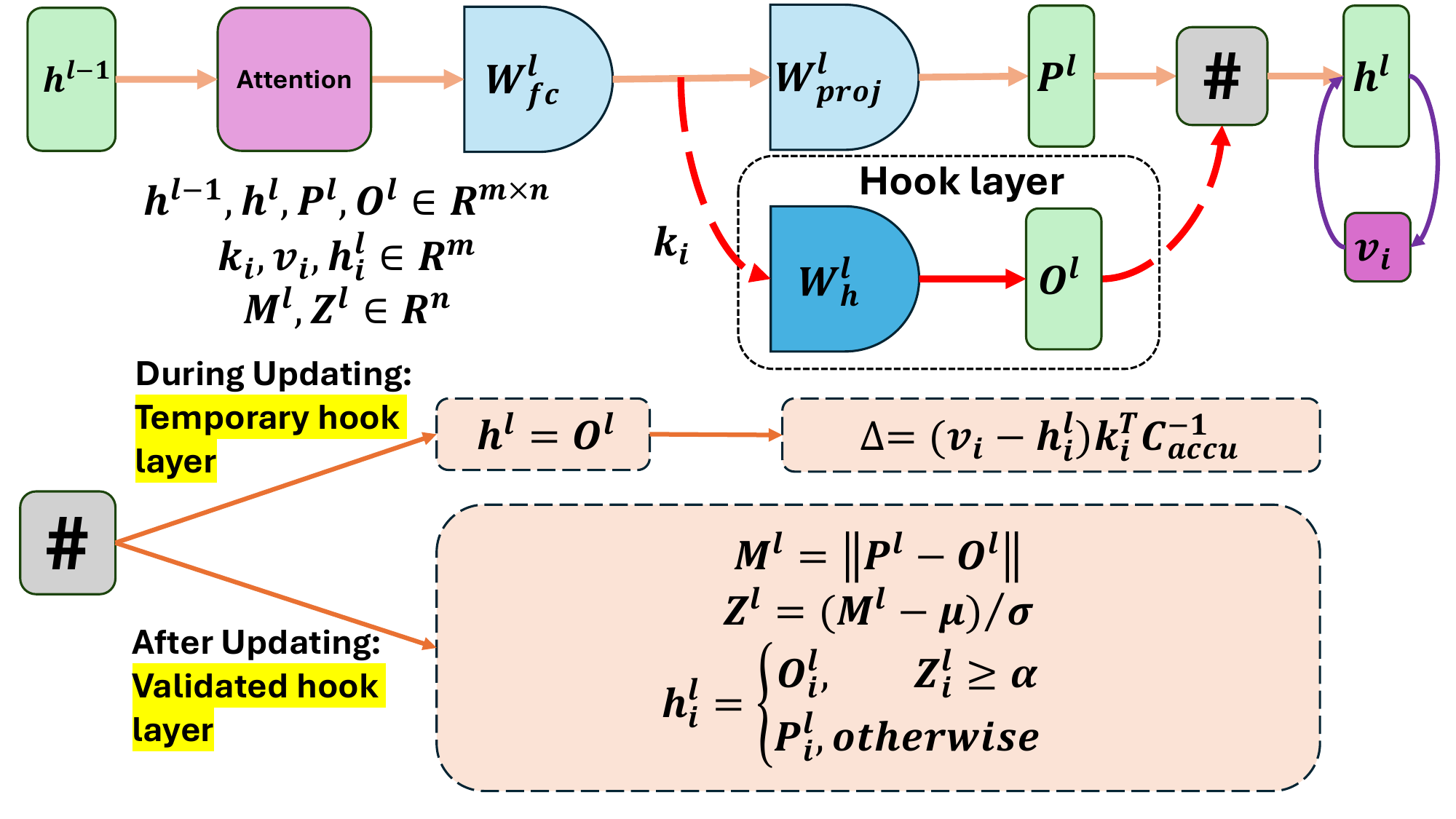}
\vspace{-0.5cm}
\caption{Single layer update with hook layer (residual connections are omitted). \(\parallel.\parallel\) means calculate the L2-norm over the keys' dimension ($m$). For each updating of a single batch edits, the temporary hook layer is used at the beginning to ensure $\Delta$ is computed based on $W_{h}^l$. After the weights update, the validated hook layer is applied to determine whether to use the original layer or hook layer for each token. This process can be implemented iteratively to support consecutive batch editing. Note that the temporary hook layer weight of a new iteration is copied from the validated hook layer weight of the previous iteration. So, the validated hook layer keeps track of the updated layer from previous edits by retaining the weight from the previous iteration.}
\label{fig. single layer update}
\vspace{-0.4cm}
\end{figure}

\subsubsection{Hook Layer}
\citet{Yao} demonstrate that those editing methods that directly modify the model parameter in place struggle with sequential editing. Specifically, the locality decreases drastically when the number of iterations increases. Meanwhile, those methods that freeze the model parameters show more stable performance over iterations. This indicates that it might be helpful to separate the editing change from the model itself. However, directly applying an external memory \cite{serac, grace} that grows over time for a consecutive batch editing scenario is too memory-costly. 
Therefore, we aim to seek an approach that could store associations without regularly increasing external memory while preserving the original model parameters.

In light of these motivations, we introduce the hook layer (Fig. \ref{fig. single layer update}), which takes the original model layer weights as the weight initialization and is responsible for all editing weight alteration in the whole editing process of CoachHooK. It is similar to the forward hook function defined in popular ML libraries like PyTorch, which adjusts the original forward layer output based on predefined criteria. Theoretically, the hook layer can be hung on any target linear layer in the transformer. Nevertheless, we mainly focus on the critical path identified in \cite{memit, rome} as they are verified to be crucial for fact association storage in the autoregressive language model.

As shown in Fig. \ref{fig. single layer update}, there are generally two sorts of hook layers in this work, namely, the \textbf{Temporary hook layer} and the \textbf{Validated hook layer}. The temporary hook layer is temporarily applied during the weight-updating process. 
It replaces the original output with the output from the hook layer so that the residual is computed based on the hook layer weight. The hook layer weights are then updated (Eq. \ref{eq.8}) using the calculated residual and the accumulated sum of the keys' outer product. Validated hook layers inherit the updated weights from the temporary hook layer, and are employed after each single-batch weight updating process at the layer.

\subsubsection{Local Editing Scope Identification}
\label{sec 2.3}
\paragraph{Outlier Detection}
Given the original outputs produced by the model layer weights and the edited outputs generated by hook layer weights, we need to decide when and which part of the original outputs to swap over.
The ideal solution is only to switch those parts of outputs whose keys have been updated to the hook layer weights and leave other parts unchanged. To this end, we first detect the output parts that have their keys updated. Suppose $k_i \in K_1, v_i \in V_1$ is an association that has been updated to $W_1$, and $k_j \notin K_1$ is a key that is not included in the updated associations. We show empirically in section \ref{sec 3.4} that $\parallel{W_1k_i-W_0k_i}\parallel \gg \parallel{W_1k_j-W_0k_j}\parallel$ holds. This implies that when the hook layer with updated weight $W_{h}$ receives an input $\hat{K} \in R^{m \times n}$ (batch dimension is ignored for simplicity) that contains an edited key $k_i \in \hat{K} \cap K_1, \, k_i \in R^m$, then we should have $\parallel{W_{h}k_i-W_{0}k_i}\parallel \gg \parallel{W_{h}k_j-W_{0}k_j}\parallel$ for all $k_j \in \hat{K}-\hat{K} \cap K_1$, which means that $\parallel{W_{h}k_i-W_{0}k_i}\parallel$ would be outliers among $\{\parallel{W_{h}k_x-W_{0}k_x}\parallel :\! \forall k_x \in \hat{K}\}$. Hence, detecting outputs of the updated keys can be transferred to detecting the outliers in the L2-norm distribution of inputs. We used the standardization to find the outliers (Fig. \ref{fig. single layer update}), which applies the standardization technique to L2-norm vectors of inputs and determines outliers via a predefined threshold $\alpha$. Concretely, for the inputs $\hat{K}$, we first compute the L2-norm vector $M^l \in R^n$:
\begin{gather}
P^l=W_{0}\hat{K}\quad O^l=W_{h}\hat{K} \\ 
M^l=\parallel(O^l-P^l)\parallel 
\end{gather}
Note that $\parallel . \parallel$ here means computing the L2-norm for each vector over the keys' dimension ($m$). Then, we standardize $M^l$ to get the z-score vector $Z^l$ and select the swap location by comparing it with $\alpha$. The details of choosing $\alpha$ are discussed in the next paragraph. Specifically, we do:
\begin{equation}
h_i^l=
    \begin{cases}
        O_i^l& \text{if  $Z_i^l \geq \alpha$},\\
        P_i^l& \text{ortherwise}.
    \end{cases}
\end{equation}
where $i$ is the index over tokens.

\paragraph{Threshold $\alpha$ Determination}
We denote $Z_{i}^l=\max((M^l-\mu)/\sigma)$ as the maximum z-score entry of an input $\hat{K}$. Since the $Z_{i}^l$ varies for different instances (\S \ref{sec 3.4}) and is likely to shift as the consecutive editing steps grow, it is unreasonable to set $\alpha$ as a fixed real number. Therefore, we determine the $\alpha$ dynamically during the editing process:
\begin{equation}
\alpha_s=
    \begin{cases}
        \alpha_z & \text{if} \; s=1, \\
        \min(\alpha_{c},\alpha_{s-1}) & \text{otherwise}
    \end{cases} 
\end{equation}
where $s\geq 1$.
Specifically, the $\alpha$ is first initialized to a pre-selected value $\alpha_z$. At each consecutive editing step $s$, for the batch of inputs in this step, we calculate $Z_{i}^l$ (the maximal z-score entry) for each single instance and select the minimal $Z_{i}^l$ in the batch (\textit{i.e.}, the supremum) as the candidate $\alpha_{c}$. The $\alpha_s$ is finally determined to be the minimum between the candidate $\alpha_{c}$ and the previous value $\alpha_{s-1}$. In practice, we set $\alpha_z=2.2$.

\begin{table*}[t]
\footnotesize
    \centering
    \setlength{\tabcolsep}{0.5mm}{
    \begin{adjustbox}{max width=\textwidth}
    \begin{tabular}{lccccccccc}
    \toprule
    \multirow{2}{*}{Method} & \multirow{2}{*}{Model} & \multicolumn{4}{c}{ZsRE} & \multicolumn{4}{c}{COUNTERFACT} \\ 
    \cmidrule(lr){3-6}\cmidrule(lr){7-10}
    
     & & Reliability & Generality & Locality & Average & Reliability & Generality & Locality & Average \\
     \midrule
     FT-L \cite{rome} & \multirow{7}{*}{GPT2-XL} & 16.85 & 16.34 & 71.55 & 34.91 & 0.27 & 0.34 & 85.18 & 28.60 \\
     FT-M && 17.95 & 17.32 & 71.26 & 35.51 & 0.36 & 0.42 & 82.81 & 27.86 \\
     LoRA \cite{LoRA} && 30.10 & 29.08 & 80.54 & 46.57 & 5.64 & 3.46 & 69.45 & 26.18 \\
     MEND \cite{mend} && 2.16 & 2.11 & 20.34 & 8.20 & 0.13 & 0.03 & 4.22 & 1.46 \\
     SERAC \cite{serac} && \textbf{98.64} & 48.12 & 35.68 & 60.81 & 17.88 & 14.55 & 82.25 & 38.23 \\
     MEMIT \cite{memit} && 61.19 & 49.97 & 97.51 & \colorbox{lime}{\textbf{69.56}} & 81.01 & 27.67 & 95.80 &  \colorbox{lime}{\textbf{68.16}} \\
     \textbf{CoachHooK} && 82.21 & \textbf{66.61} & \textbf{99.40} & \colorbox{green}{\textbf{82.74}} & \textbf{88.28} & \textbf{40.38} & \textbf{97.66} &  \colorbox{green}{\textbf{75.44}} \\
     \midrule
     FT-L \cite{rome} & \multirow{6}{*}{GPT-J} & 22.57 & 21.77 & \textbf{99.19} & 47.84 & 0.37 & 0.34 & 99.57 & 33.43 \\
     FT-M             && 99.96 & 80.31 & 43.35 & 74.54 & \textbf{99.99} & 35.29 & 17.04 & 50.77 \\
     LoRA \cite{LoRA} && \textbf{99.97} & \textbf{83.20} & 17.64 & 66.93 & 99.87 & \textbf{53.10} & 2.50 & 51.82 \\
     SERAC \cite{serac} && 87.46 & 63.64 & 77.35 & 76.15 & 16.67 & 15.93 & \textbf{99.99} & 44.20 \\
     MEMIT \cite{memit} && 93.40 & 70.45 & 96.47 & \colorbox{lime}{\textbf{86.77}} & 99.57 & 42.29 & 95.25 & \colorbox{green}{\textbf{79.04}} \\
     \textbf{CoachHooK} && 97.59 & 72.41 & 99.10 & \colorbox{green}{\textbf{89.70}} & 87.94 & 42.76 & 98.17 & \colorbox{lime}{\textbf{76.29}} \\     
    \bottomrule
    \end{tabular}
    \end{adjustbox}}
    \caption{Single round batch editing results. The best two average scores are highlighted. All metrics are desired higher values.}
    \label{tab:single round batch editing}
    \vspace{-0.3cm}
\end{table*}

\begin{figure}
\centering
\includegraphics[width=0.48\textwidth]{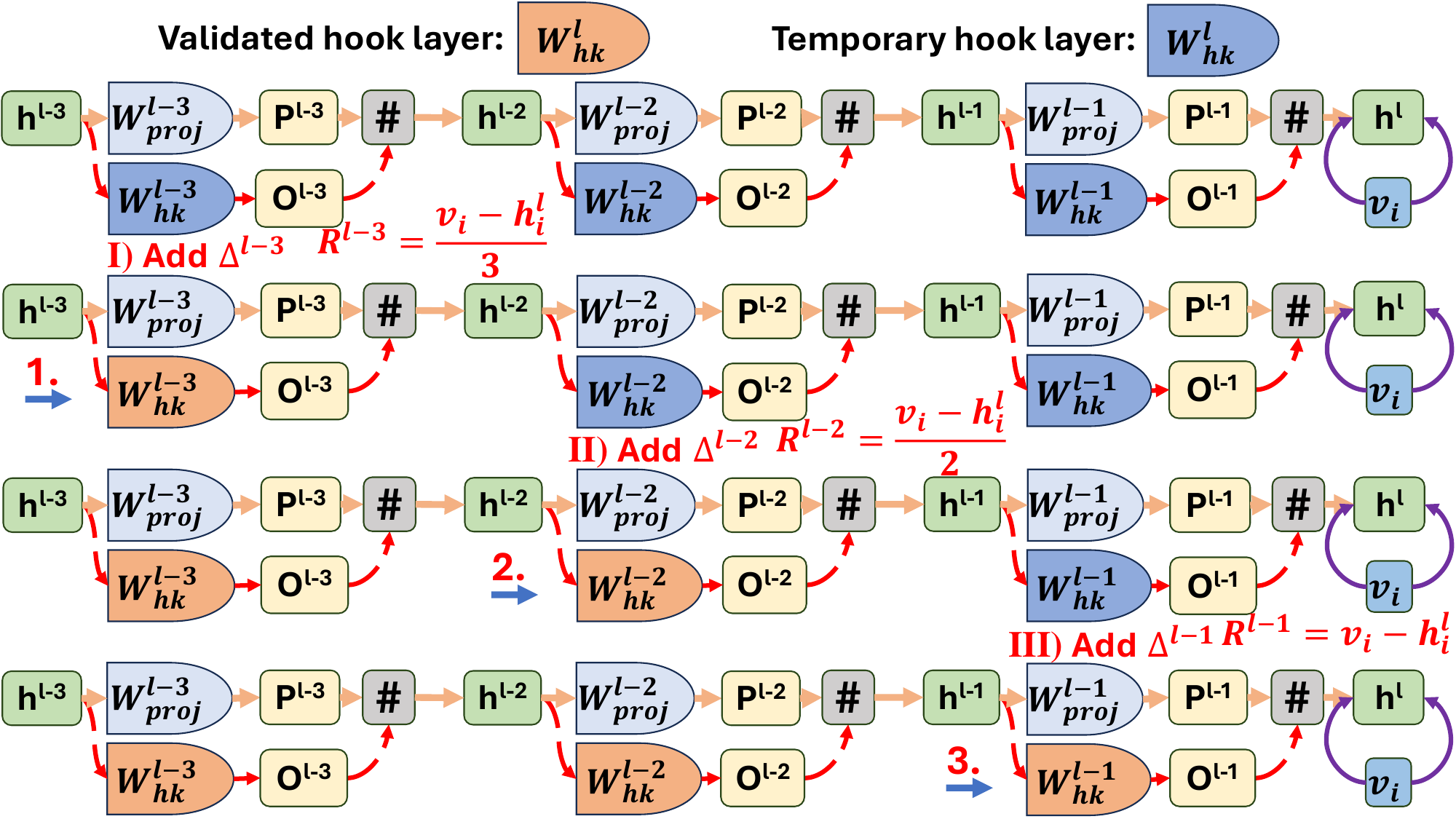}
\vspace{-0.5cm}
\caption{Multiple layer update with hook layer (Attention module and the first layer of FFN are omitted). The value vector $v_i$ is first computed at the last editing layer, and then we iteratively insert a fraction of the residual to each editing layer (I, II, III) using Eq. \ref{eq.8}. Since changing one layer would affect the activations of downstream layers, recollection of the activations is conducted after each iteration. At the beginning, temporary hook layers are initialized to all editing layers. Once the hook layer weight is updated, it is replaced by the validated hook layer (1, 2, 3).}
\label{fig. Multi-layer update}
\vspace{-0.5cm}
\end{figure}

\subsection{Multiple-layer Consecutive Batch Editing}
\label{sec 3.3}

Given the designed single-layer editing procedure, there exists a risk that the single-layer hook fails to detect the updated keys.
Suppose $k_i$ is an updated key; failure to detect $k_i$ indicates that the output corresponds to $k_i$ at this single layer would be the same as the original output $W_{proj}k_i$, which consequently leads to the failure update for $k_i$. 
To tackle this issue, one potential solution is to apply the hook to multiple model layers rather than a single model layer because the latter layer grasps the chance to capture the edited keys missed by proceeding layers. Furthermore, \cite{modify_memory} showcased that minimizing the magnitude of parameter change is helpful for improving the robustness of the model. Thus, we expand our work to multiple layers (Fig. \ref{fig. Multi-layer update}).

We first find the desired object vector $v_i$ following a similar procedure in \cite{memit}. However, the optimization is not based on the original model, but the model hung with the validated hook that inherits the most recently updated hook weights from the previous editing step. After $v_i$ is found, 
the hook weight is updated at each layer.

At each batch editing step, all the hook layers are initialized to temporary hook layers, which substitute the entire original output to output from hook layers. The purpose of doing this is to ensure that the residual regarding the hook layer weights rather than the original model weights are calculated. Then, the residual is distributed evenly to each layer, and the alteration $\Delta^l$ to the parameter at each layer is found in a layer-increasing iterative manner with keys and residuals recomputed at each iteration (Fig. \ref{fig. Multi-layer update}). The reason for the recomputation of keys and residuals is that the layer-increasing alteration approach will affect the keys and residuals in the latter layer. For each layer, once the hook layer weight is updated, the hook layer is changed from a temporary hook layer to a validated hook layer to facilitate the computation of the keys and residuals in the latter layer. After the whole editing process is completed, the validated hook layers with the ultimately updated weights are hung on the model to shape the final edited model. 

\section{Experiments}
\subsection{Experiment Setups}
\label{sec:experiment setup}

\paragraph{Datasets \& Evaluation Metrics}
We use the ZsRE \cite{zsre} and COUNTERFACT \cite{memit} datasets with the split provided in EasyEdit\footnote{\url{https://github.com/zjunlp/EasyEdit/tree/main}} for evaluation. We employ three popular editing evaluation metrics defined in \cite{Yao, patcher, decao}, \textit{i.e.}, Reliability, Generality, and Locality, as well as the average scores\footnote{Most of the application scenarios of model editing require good performance in all three metrics.} over the three metrics. Further details are provided in Appendix \ref{appendix_A}. 

\paragraph{Baselines \& Implementation Details}
For baselines
, we adopt several batch-supportive editing methods, including LoRA \cite{LoRA}, SERAC \cite{serac}, MEND \cite{mend}, MEMIT \cite{memit} and fine-tuning with specific layer (FT-L) technique used in \cite{rome, Yao}, which only fine-tune a specific layer identified by Rome \cite{rome} instead of all layers to ensure a fair comparison. We also include a small variation of FT-L called FT-M and a sequential supportive editing method GRACE \cite{grace}. We choose large autoregressive language models GPT2-XL and GPT-J (6B) as our base models.  

For all consecutive editing experiments, the evaluation is conducted after the full set of consecutive steps are finished. For example, in Fig. \ref{fig. wohk}, we conduct experiments for sample sizes 200, 400, 600, 800, and 1000, so the evaluation is triggered right after the first 200, 400, 600, etc, samples are edited to the model. Unless specified, the batch size\footnote{Since GRACE \cite{grace} does not support batch editing, we set the batch size to 1 for GRACE.} for consecutive editing is selected to be 10. Further details of the baselines and implementation are given in the Appendix \ref{app:baseline}.

\begin{table*}[t]
\footnotesize
    \centering
    \setlength{\tabcolsep}{0.5mm}{
    \begin{adjustbox}{max width=\textwidth}
    \begin{tabular}{lccccccccc}
    \toprule
    \multirow{2}{*}{Method} & \multirow{2}{*}{Model} & \multicolumn{4}{c}{ZsRE} & \multicolumn{4}{c}{COUNTERFACT} \\ 
    \cmidrule(lr){3-6}\cmidrule(lr){7-10}
    
     & & Reliability & Generality & Locality & Average & Reliability & Generality & Locality & Average \\
     \midrule
     FT-L \cite{rome} & \multirow{8}{*}{GPT2-XL} & 3.79 & 2.48 & 6.60 & 4.29 & 1.00 & 1.00 & 6.00 & 2.67 \\
     FT-M             && 8.92 & 8.41 & 6.22 & 7.85 & 4.00 & 3.50 & 5.50 & 4.33 \\
     LoRA \cite{LoRA} && 0.96 & 1.29 & 0.03 & 0.76 & 0.50 & 0.02 & 0.50 & 0.34 \\
     MEND \cite{mend} && 20.95 & 18.29 & 93.69 & 47.01 & 0.01 & 0.00 & 0.08 & 0.03 \\
     SERAC \cite{serac} && 100 & 36.03 & 35.95 & 57.33 & 15.41 & 12.96 & 81.00 & 36.46 \\
     GRACE \cite{grace} && \textbf{100} & 0.04 & \textbf{100} & \colorbox{lime}{\textbf{66.68}} & \textbf{100} & 0.40 & \textbf{100} & \colorbox{green}{\textbf{66.80}} \\
     MEMIT \cite{memit} && 34.88 & 32.96 & 70.74 & 46.19 & 56.00 & 37.00 & 31.00 & 41.33 \\
     \textbf{CoachHooK} && 66.91 & \textbf{56.11} & 97.23 & \colorbox{green}{\textbf{73.42}} & 86.00 & \textbf{38.00} & 59.00 & \colorbox{lime}{\textbf{61.00}} \\
     \midrule
     FT-L \cite{rome} & \multirow{7}{*}{GPT-J} & 23.53 & 21.70 & 55.27 & 33.5 & 2.00 & 2.00 & 72.00 & 25.33 \\
     FT-M             && 64.33 & 55.63 & 17.59 & 45.85 & 25.50 & 5.00 & 2.00 & 10.83 \\
     LoRA \cite{LoRA} && 1.43 & 1.39 & 0.02 & 0.95 & 0.50 & 0.50 & 0.10 & 0.37 \\
     SERAC \cite{serac} && 86.91 & 55.36 & 79.07 & 73.78 & 18.49 & 14.56 & 98.89 & 43.98 \\
     GRACE \cite{grace} && \textbf{100} & 0.04 & \textbf{100} & \colorbox{lime}{\textbf{66.68}} & \textbf{100} & 0.50 & \textbf{100} & \colorbox{lime}{\textbf{66.83}} \\
     MEMIT \cite{memit} && 63.36 & 48.90 & 74.80 & 62.35 & 75.00 & \textbf{45.00} & 42.00 & 54.00 \\
     \textbf{CoachHooK} && 79.89 & \textbf{61.29} & 96.52 & \colorbox{green}{\textbf{79.23}} & 95.00 & 41.00 & 80.00 & \colorbox{green}{\textbf{72.00}} \\
    \bottomrule
    \end{tabular}
    \end{adjustbox}}
    \caption{Consecutive batch editing results.}
    \label{tab:consecutive batch editing}
    \vspace{-0.4cm}
\end{table*}

\subsection{Evaluation on Single-round Batch Editing}

We first test the effectiveness of our method under basic single-round batch editing settings with batch size 30, \textit{i.e.}, the model is rolled back to the initial state after each batch editing. Both MEMIT and CoachHooK need to set the parameter $\lambda$, the balance factor between pre-trained and newly updated associations. According to \cite{memit}, higher $\lambda$ helps preserve the original model behavior (locality) but could harm reliability and generality, and the best overall performance is found at around $\lambda = 10^4$. However, with the intent to verify whether our method comprehensively improves the editing, that is, could accept lower $\lambda$ to assign higher weight for new associations while not sacrificing the locality, we deliberately set $\lambda=5\times10^3$ for CoachHooK and keep it as the optimized value for MEMIT, which are $2\times10^4$ and $1.5\times10^4$ for GPT2-XL and GPT-J respectively.

The evaluation results are shown in Table \ref{tab:single round batch editing}. For GPT2-XL, our method has the best result in almost every metric. Specifically, despite the relatively low $\lambda$, our method overwhelms other baselines in generality metrics while maintaining a better locality. This indicates that lowering $\lambda$ or, in other words, increasing the weight of the new associations does not sacrifice the locality in CoachHooK. The improvement in GPT-J is less compared with that in GPT2-XL. However, our method still has the best average score for the ZsRE dataset and a comparable average score with the best in the COUNTERFACT dataset.

\subsection{Evaluation on Consecutive Batch Editing}

We evaluate our method's capability on 1k samples from both datasets for consecutive batch editing, \textit{i.e.}, there is no roll-back. The evaluation is conducted after the end of the whole consecutive batch editing process. We set $\lambda$ to $15,000$ as the scenario now is consecutive batch editing. 

Results in Table \ref{tab:consecutive batch editing} show that most of the methods suffer from a great performance drop contrasted to editing in a single round. Although our method's performance experiences a decrease as well, it surpasses other methods in 100 consecutive steps with an even larger improvement margin for almost all the metrics compared to the single-round batch editing. This demonstrates that our method does not depend on simple regurgitation of the editing samples nor rely heavily on the trade-offs of lowering the balancing factor $\lambda$ to increase the reliability and locality. An interesting point is that the GRACE performs perfectly in reliability and locality but poorly in generality. 
As expected, GRACE is superior in reliability since it maintains a codebook to memorize the instances of editing that are encountered. However, its inferiority in generality indicates that it suffers from the problem of regurgitation. (One may argue that increasing the deferral radius could help improve the generality of GRACE, we therefore provide further experiments in Appendix \ref{appendix_C}.)

We extend the data scale of the consecutive batch editing experiment to 10k (1k consecutive steps) to explore the limit of our method. Results can be found in Fig. \ref{fig:namada 10k}. Surprisingly, the locality experiences a great fall from 100 to 200 steps but remains steady from 200 to 1k editing steps, which proves that the hook layer stably obstructs the out-scope samples. Reliability and generality consistently fall as the consecutive steps grow, indicating that there is still room for improvement in this field.

\subsection{Validation of Local Editing Scope}
\label{sec 3.4}
\begin{figure}[t]
\centering
\includegraphics[width=0.48\textwidth]{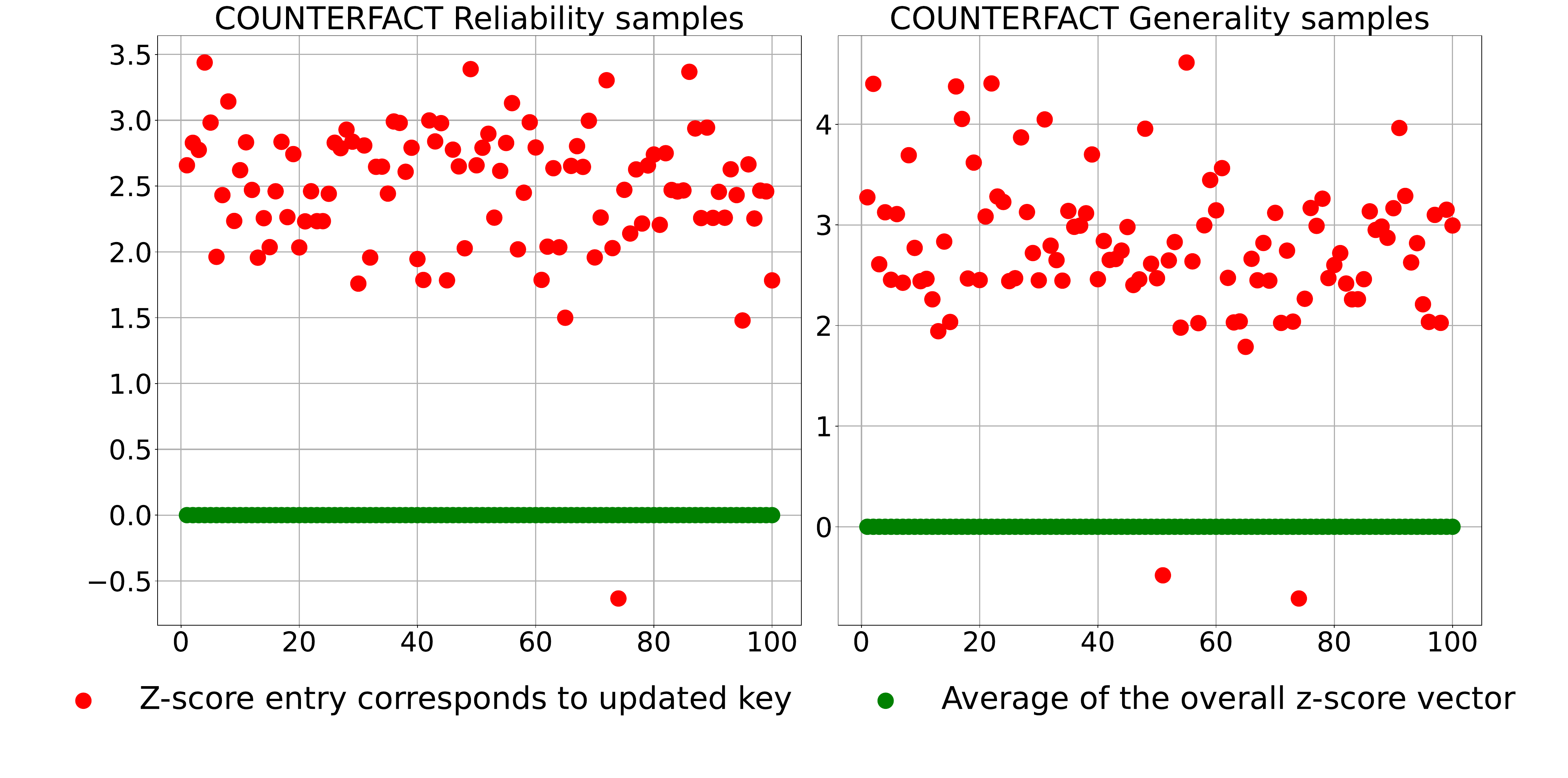}
\vspace{-0.6cm}
\caption{Difference between the z-score entry to the updated key $Z^l_{key}$ and average of $Z^l$. The x-axis represents the sample index.}
\label{fig:editing_scope}
\vspace{-0.6cm}
\end{figure}

Given an updated hook layer with the weight $W_{h}$, the original model weight $W_{0}$, an updated key $k_i$, and an out-of-scope key $k_j$, we conduct experiments to verify whether $\parallel{W_{h}k_i-W_{0}k_i}\parallel\gg \parallel{W_{h}k_j-W_{0}k_j}\parallel$ holds.
We select 100 samples from the COUNTERFACT dataset to edit GPT2-XL using CoachHooK, then apply the edited model to these 100 samples and record the z-score entries of the L2-norm of the difference vector between update keys' response from the last hook layer and original model layer, namely, z-score entries of $\parallel{W_{h}k_i-W_{0}k_i}\parallel$. Both reliability and generality prompts are included for comprehensiveness.

The result is shown in Fig. \ref{fig:editing_scope}. Almost all the z-scores of the responses from updated keys exhibit a great margin from the mean value, with the lowest around 1.5 in reliability samples and 2 in generality samples. 
The discriminative z-score demonstrates that the identification technique (section \ref{sec 2.3}) can effectively filter editing-irrelevant instances and accept editing-relevant instances, which validates the local editing scope.

\subsection{Detailed Analysis and Discussions}
\begin{figure}[t]
\setlength{\abovecaptionskip}{0pt}   
\setlength{\belowcaptionskip}{0pt}
\centering
\includegraphics[width=0.48\textwidth]{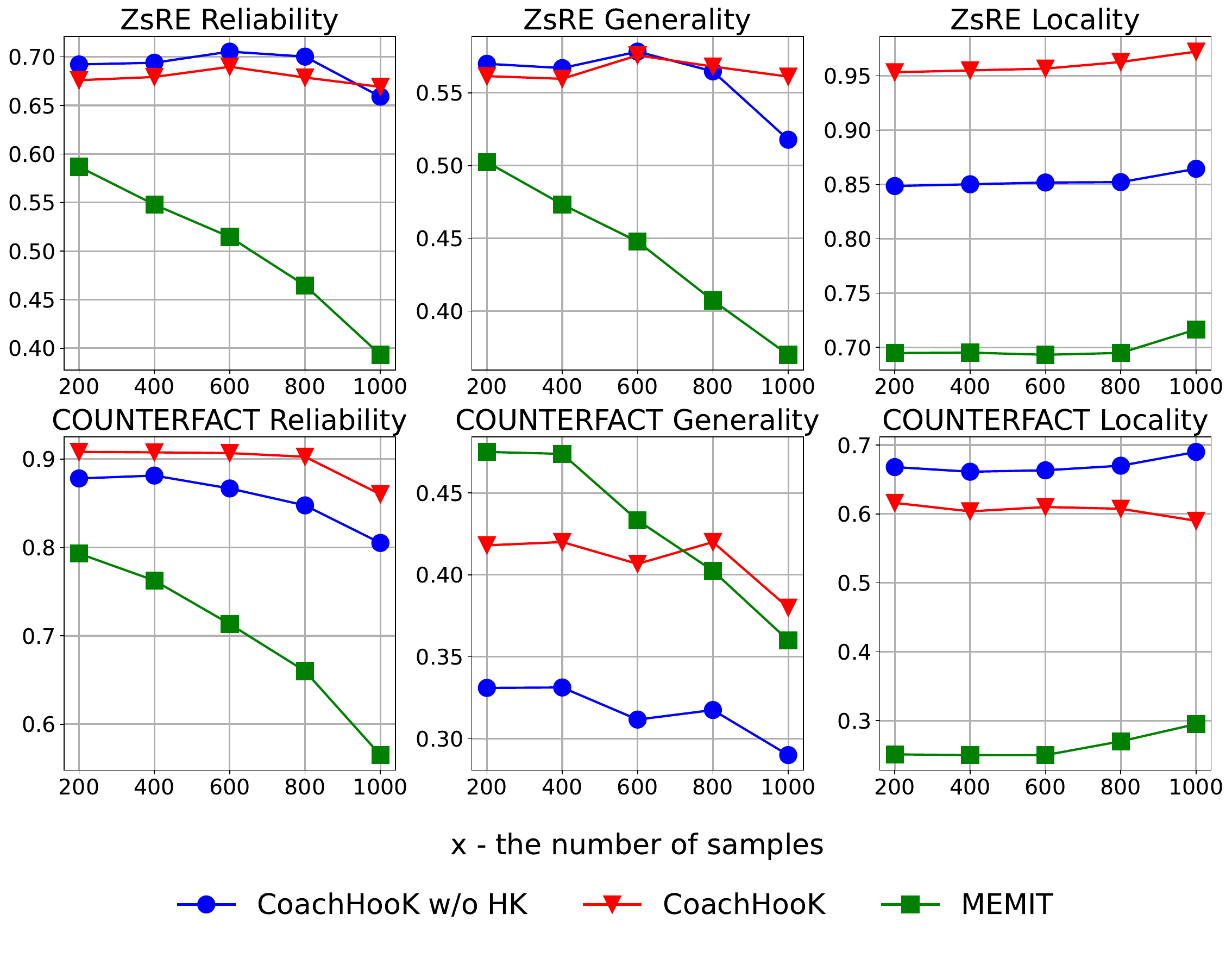}
\vspace{-0.6cm}
\caption{Ablation study
.}
\label{fig. wohk}
\vspace{-0.6cm}
\end{figure}

\paragraph{Ablation Study of Update mechanism and Hook layers}
The effectiveness of the derived consecutive updating mechanism and the hook layers are discussed in this part. We run three cases using GPT2-XL, namely, MEMIT (no consecutive updating mechanism, no hook layers), CoachHooK without hook (CoachHooK w/o HK), and CoachHooK for consecutive batch editing on 1k samples from both ZsRE and COUNTERFACT datasets. 

The results are demonstrated in Fig. \ref{fig. wohk}. In almost all metrics of the two datasets except the generality of COUNTERFACT, the CoachHooK w/o the hook performs better than the vanilla MEMIT, and the margin tends to increase as the consecutive steps ascend. This certifies the effectiveness of our derived consecutive updating mechanism in consecutive batch editing scenarios. For the ZsRE dataset, the method with hook layers considerably outperforms the one without hook in the locality without sacrificing reliability and generality. This verifies that the hook layer can efficiently and accurately block the out-scope instances from the input without fraudulently missing in-scope instances. For the COUNTERFACT dataset, the reliability of CoachHooK is consistently higher than the other two, and the generality surpasses that of MEMIT after 80 editing steps. Besides, the hook layer causes some side effects in the locality of COUNTERFACT, but this circumstance is not found in the ZsRE dataset. It is worth noting that CoachHooK shows the most stable performance as the number of consecutive editing steps grows, showing the great potential of our method for consecutive editing.

\paragraph{Effect of the Balance Factor \(\lambda\)}
\label{vary_lamada}
\begin{figure}
\centering
\includegraphics[width=0.48\textwidth]{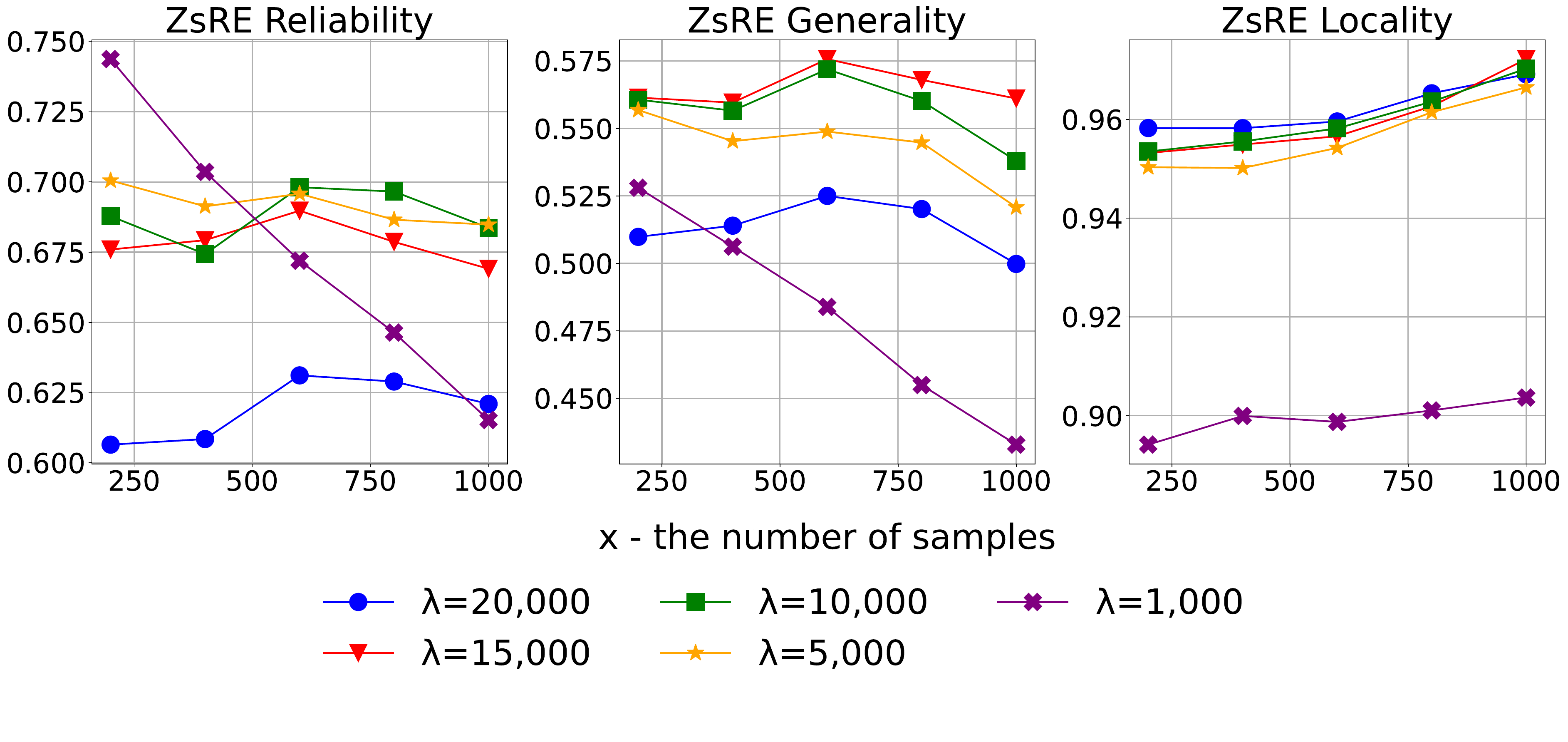}
\vspace{-1cm}
\caption{Performance comparisons on initial five different values of $\lambda$.}
\label{fig:namada 1k}
\vspace{-0.3cm}
\end{figure}

\begin{figure}
\centering
\includegraphics[width=0.48\textwidth]{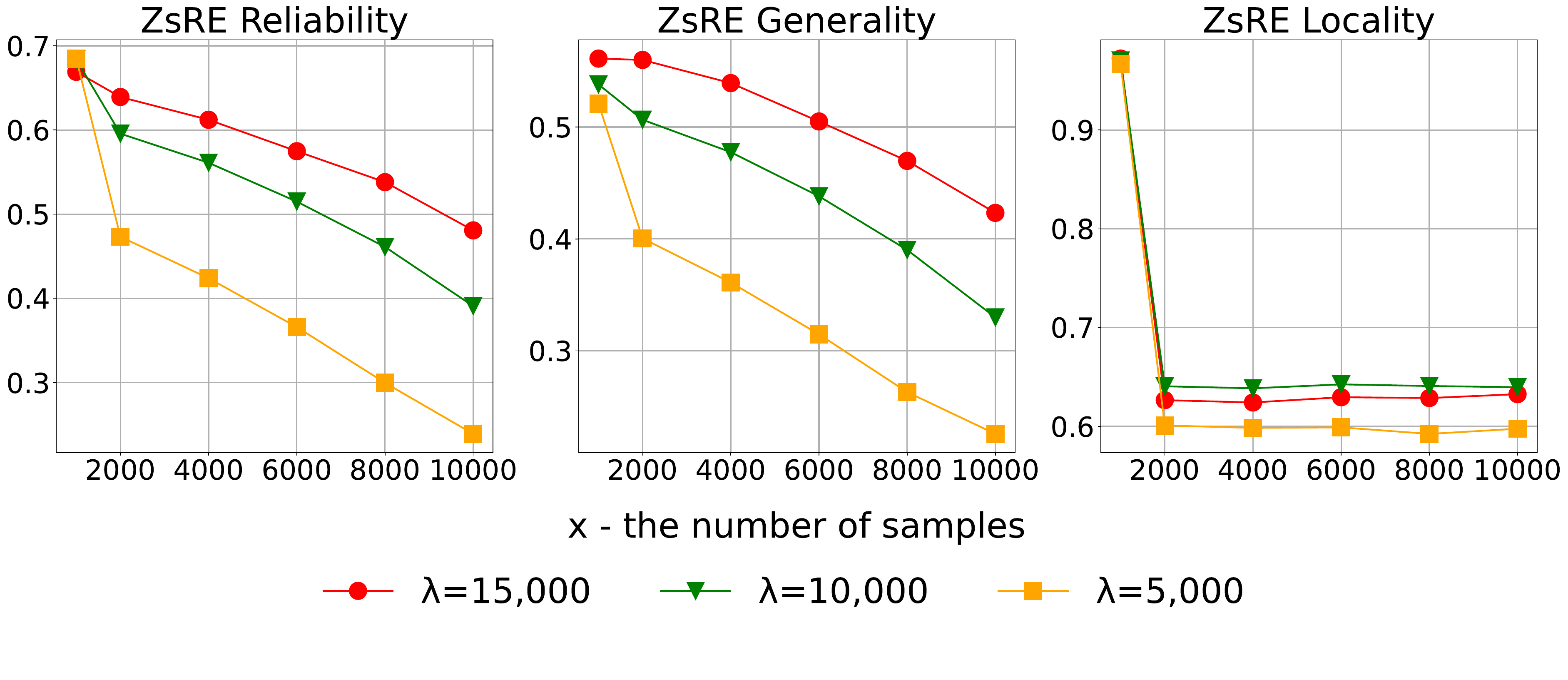}
\vspace{-1cm}
\caption{Extension on the best three values of $\lambda$.}
\label{fig:namada 10k}
\vspace{-0.6cm}
\end{figure}

We test the effect of different $\lambda$, the balance factor between pre-training and newly updated associations. We first evaluate the CoachHooK with different $\lambda$ on 1k samples from ZsRE (Fig. \ref{fig:namada 1k}). It seems that a small value of $\lambda=1,000$ would cause significant damage to all three metrics, especially the reliability and generality, since they experience a great drop as the consecutive steps increase. This may result from the overly high magnitude of the weight change caused by the low value of $\lambda$, which severely distorts the previously updated associations. Meanwhile, a too-high value of $\lambda=20,000$ also seems not to be a good choice, which gives rise to an overly small magnitude of the weight change so that it fails to deliver the new optimized values for keys. The cases of $\lambda=5000, 10000, 15000$ do not show an apparent difference, so we extend further the sample size to 10k (Fig. \ref{fig:namada 10k}).

Extended results show that 5000 is not a good choice for large-consecutive editing steps, though it performs no worse than the other two in early 1k samples. The case of $\lambda=15,000$ ranks first in reliability and generality. Although it performs worse in locality compared to $\lambda=10,000$, the margin between them gradually narrows as the consecutive steps rise. Overall, we conclude that 15,000 would be a reasonable selection.

\paragraph{Effect of Editing Batch Size}
\begin{figure}
\centering
\includegraphics[width=0.48\textwidth]{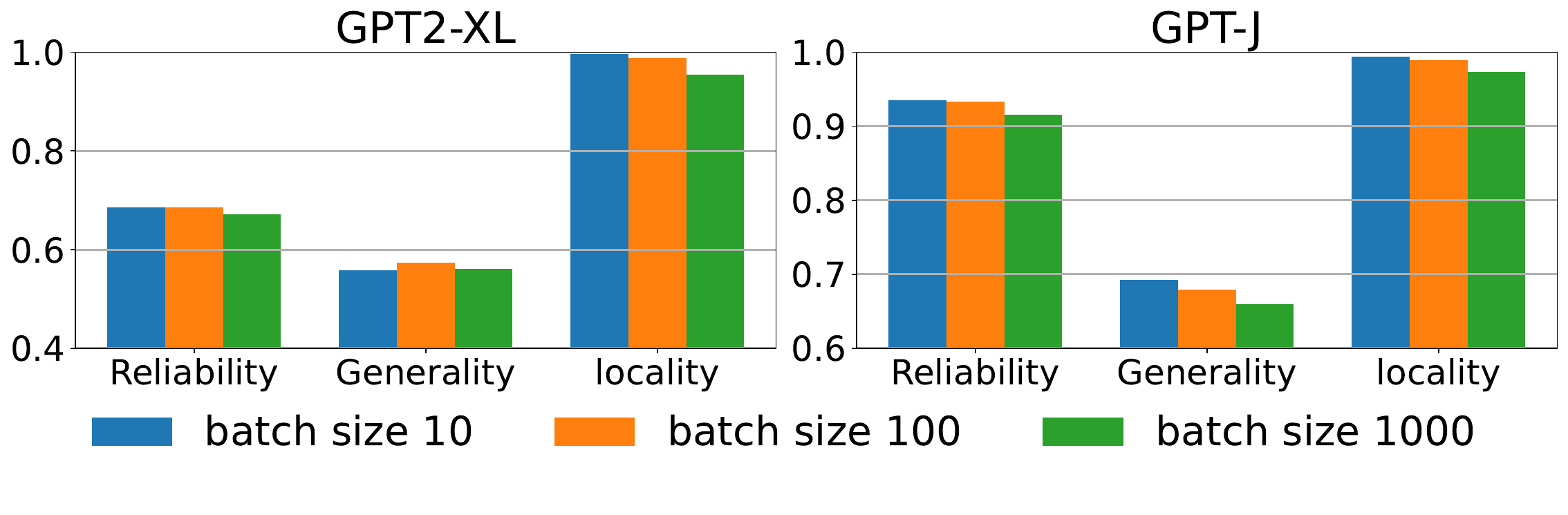}
\vspace{-0.8cm}
\caption{Performance comparisons on different editing batch sizes.}
\label{fig:vary_batch_size}
\vspace{-0.5cm}
\end{figure}
Does the batch size parameter affect the performance of our method? We investigate the effect of batch size by conducting single-round editing on 1k samples from ZsRE. We tested batch sizes 10, 100, and 1000 (Fig. \ref{fig:vary_batch_size}).

The results show that while the three metrics decrease as the batch size rises, the margin could be negligible, denoting that our method possesses the mass-editing capacity.

\paragraph{More Analyses} Other detailed analyses including hyperparameters search, inference time and memory analysis are presented in Appendix \ref{appendix_C}.

\section{Related Work}

\paragraph{Linear Associative Memory}
Linear Associative Memory \cite{linear_memory, linear_memory1} treats the model weights $W$ as a set of associations between a set of keys and values. Many researchers have adopted this technique as a memory component in their methods. \cite{thirdorder, stepgame, depwignn, mapping, fast_weight}. Recent research \cite{trans-ffn} finds that the transformer feed-forward layers are likely to be a key-value memory. This significantly helps researchers to understand what is happening inside the transformer black box and inspires researchers in the field of model editing to first locate where the knowledge is stored and then edit the corresponding parameters \cite{memit, rome}.

\paragraph{Model Editing}
Recent years have witnessed prosperous development in the field of model editing. According to \cite{Yao}, the proposed methods so far can be generally classified into two groups, \textit{i.e.}, modify the model's weight or not. 
The methods that do not directly alter the model weights generally follow two directions: they either employ an external memory or introduce additional adjustable parameters. Methods like T-Patcher \cite{patcher} and CaliNET \cite{calinet} apply new neurons that are responsible for specific mistakes in the last layer of the FFN model. GRACE \cite{grace} introduces a timely adjusted code book to edit the model's behavior. Another group of methods like SERAC \cite{serac} integrates an explicit external memory as edit descriptors to help editing scope recognition. Some other techniques like IKE \cite{ike}, MeLLo \cite{mello}, and MemPrompt \cite{memprompt} take advantage of the recent development of in-context learning and use carefully designed demonstration context prompts to edit the model. \citet{on-the-transformation} also discussed the possible transformation between context and parameter update. 
On the other hand, those directly altering the model's weight either train a hyper-network to predict the change required by the edits or first locate corresponding parameters responsible for specific knowledge and then edit the located parameters. For example, Methods like KE \cite{ke}, MEND \cite{mend}, and MALMEN \cite{malmen} follow the idea of meta-learning and use the trained hypernetwork to predict the desired parameter change for editing. ROME \cite{rome}, MEMIT \cite{memit}, PMET \cite{pmet}, KN \cite{kn}, and WilKE \cite{wilke} first locate the parameters responsible for knowledge and then using a designed optimization mechanism to shift the parameters.

Despite the good performance of the proposed methods, researchers investigated the harmful consequences of these methods. \citet{Butterfly-Effect} points out even few edits could result in the collapse of a model. \citet{Catastrophic-Forgetting} shows that scaling up the edits could lead to catastrophic forgetting of the previous edits. \citet{hurt-general-abilities} argues that editing could lead to the degradation of the model's general capabilities. Although the rationality of these investigations is still under discussion, they reveal the potential risk that model editing techniques could bring to the base model.

\section{Conclusion}
This work introduces a novel model editing method, CoachHooK, which advocates the more practical consecutive batch model editing. CoachHooK uses an expanded 
editing mechanism to support consecutive editing and newly proposed hook layers to identify the editing scope. Compared to existing model editing methods, CoachHooK does not require large external memory nor extra training for meta-networks or classifiers. 
Instead, it adopts hook layers whose size remains fixed over time for storing associations. Comprehensive experiments are conducted to verify the method's effectiveness over single-round and consecutive batch editing. 

\section*{Limitations}
Several aspects remain to be further investigated.

\paragraph{Other types of tasks}
Notably, model editing techniques could be applied to various types of tasks. Specifically, besides factual knowledge editing, it can be applied to erase hallucinations, biases, privacy information, etc. However, the concentration of this paper is to explore the practicability of expanding the model editing application scenario to consecutive batch editing and investigate the potential bottleneck of corresponding methods under this scenario. Therefore, our experiment focuses on varying the scale of editing samples in factual knowledge editing tasks, as it is a relatively well-studied and universal evaluation task in model editing.

\paragraph{Model scale and architecture}
Due to the limited computational resources, we cannot verify our method's effectiveness in larger LLMs such as Llama-2 \cite{llama2}, and GPT-NEOX-20B \cite{gpt-neox-20b}. We focus on the decoder-only autoregressive models and do not include encoder-decoder structure models, as the autoregressive structures are the mainstream architecture nowadays \cite{gpt-4, llama2}. Further, as stated by \citet{Yao}, the weight matrix in some models like OPT-13B \cite{opt-13b} is not invertible. However, 
such an issue can be relieved by adding a term $\beta I$ to the Eq. \ref{eq.14}, where $\beta$ is a scalar expected to be small and $I$ is the identity matrix.

\paragraph{The shrink of $\alpha$}
As more and more associations are integrated into the hook layer, the dynamically determined hyperparameter $\alpha$ will gradually shrink, meaning that an increasing number of vector entries in the original layer output will be swapped by the output from the hook layer, which is likely to lead the drop in locality. Nevertheless, such a problem can be alleviated by the newly designed updated mechanism (Eq. \ref{eq.8}), which considers both previously updated and newly updated keys.

\bibliography{custom}

\appendix

\section{Experiment Details}
\label{appendix_A}
All baselines are implemented using the EasyEdit \cite{EasyEdit} library. 

\paragraph{Evaluation Metrics}
We employ three popular editing evaluation metrics defined in \cite{Yao, patcher, decao}, \textit{i.e.}, reliability, generality, and locality. Given an initial base model $f_{\theta}$, a post-edit model $f_{\theta'}$, and a set of edit instances $\{(x_t,y_t)\}$, the reliability is computed as the average accuracy of the edit cases:
\begin{equation}
\mathbb{E}_{(x_t,y_t)\in \{(x_t,y_t)\}} \{ \arg\max\nolimits_y f_{\theta'}(y|x_t) = y_t \} \ .
\end{equation}
The editing should also edit the equivalent neighbor of the instance $N(x_t,y_t)$ (\textit{e.g.} rephrased descriptions). This metric is named generality and is evaluated by the average accuracy on the neighbors of the edit cases:
\begin{equation}
\mathbb{E}_{(x_t',y_t')\in \{N(x_t,y_t)\}} \{ \arg\max\nolimits_y f_{\theta'}(y|x_t') = y_t' \} \ .
\end{equation}
Despite the editing, those instances that are irrelevant to the edit cases $\{O(x_t,y_t)\}$ should not be affected. This evaluation is called locality (also known as specificity) and is measured by the proportion of unchanged predictions between the initial model and the post-edit model:
\begin{equation}
\mathbb{E}_{(x_t',y_t')\in \{O(x_t,y_t)\}} \{ f_{\theta'}(x_t') = f_{\theta}(x_t') \} \ .
\end{equation}

\paragraph{Datasets}

\begin{figure}
\centering
\includegraphics[width=0.48\textwidth]{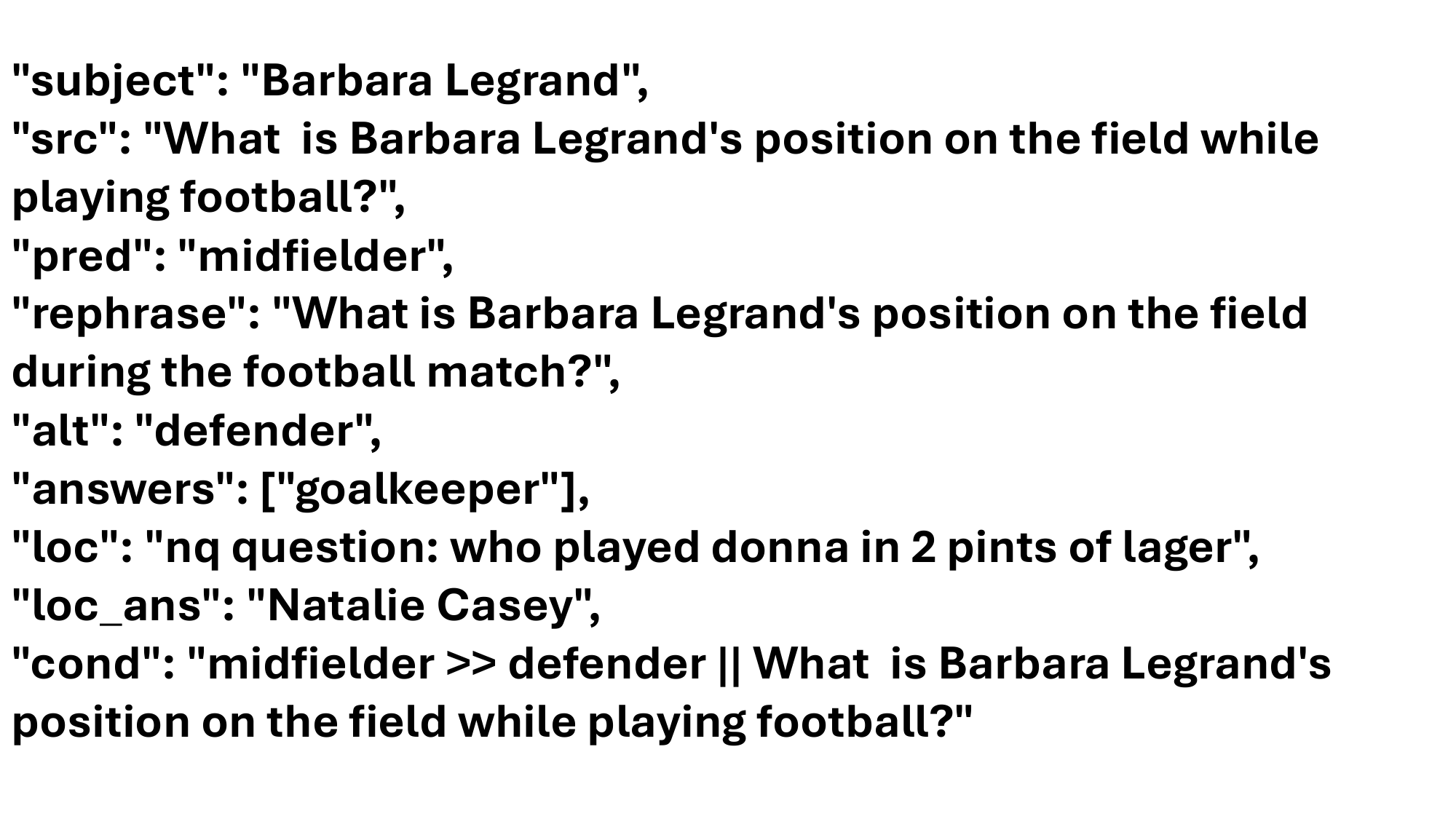}
\vspace{-0.8cm}
\caption{A sample from ZsRE dataset.}
\label{fig:ZsRE sample}
\vspace{-0.3cm}
\end{figure}

\begin{figure}
\centering
\includegraphics[width=0.48\textwidth]{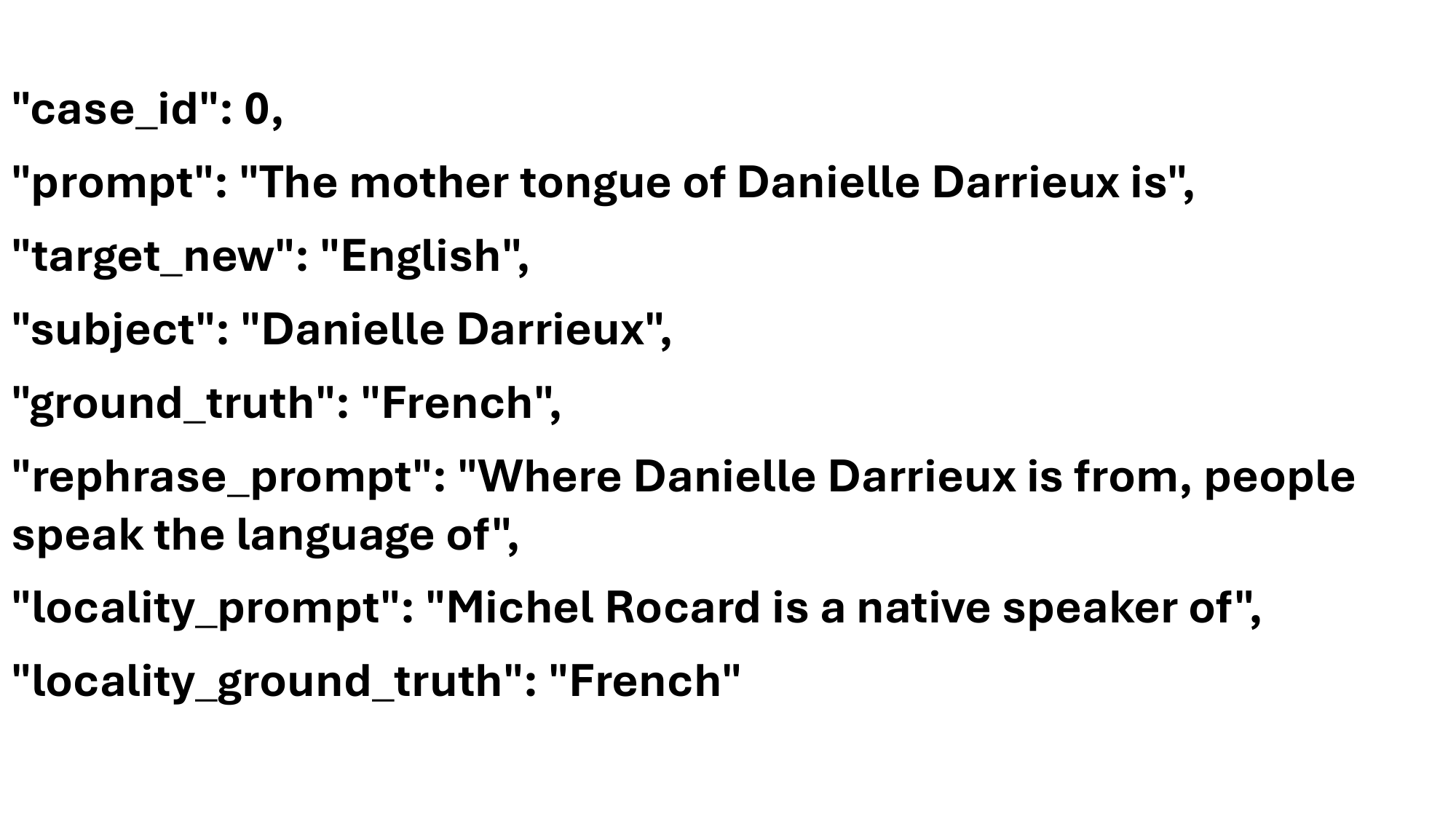}
\vspace{-0.8cm}
\caption{A sample from COUNTERFACT dataset.}
\label{fig:Fact sample}
\vspace{-0.3cm}
\end{figure}

ZsRE is a question-answering dataset that uses back-translation to generate equivalent neighborhoods. It is initially used in factual knowledge evaluation and later adopted in model editing by \cite{mend}. COUNTERFACT is a challenging dataset focusing on counterfactual information with a low prediction score compared to correct facts. It builds out-of-scope data by replacing the subject entity with a similar description that shares the same predicate.
 
Fig. \ref{fig:ZsRE sample} shows an example from the ZsRE dataset. Each record in ZsRE contains the subject string, the factual prompt used for testing reliability, the rephrase prompt used for generality evaluation, and the locality prompt used for evaluating the locality. Note that what the locality demands the post-edit model does is not to predict the ground truth but whatever the initial base model predicts. Similarly, the fact, rephrase, and locality prompts of each record in COUNTERFACT are applied to the evaluation of the three metrics respectively (Fig. \ref{fig:Fact sample}). 

\section{Baselines and Implementation Details}\label{app:baseline}
\paragraph{Fine-tuning}
We implemented two fine-tuning methods in the experiments. For FT-L, we follow the procedure in \cite{memit, rome} and fine-tune the $mlp_{proj}$ parameter from layer 0 for GPT2-XL and layer 21 for GPT-J since they are found to have the optimal performance. FT-M\footnote{\url{https://github.com/zjunlp/EasyEdit/blob/main/hparams/FT/gpt2-xl.yaml}} is a small variation of FT-L, and it uses a different loss computation procedure to optimize the parameters. For both models, we conduct 25 optimization steps using Adam optimizer \cite{adam} and use learning rate $5e^{-4}$. All other parameters of both models follow default settings.

\paragraph{LoRA}
\citet{LoRA} proposed a parameter-efficient fine-tuning method that decomposes the update gradient matrix into two small rank-n matrices, which reduces the required memory for LLM training to a large extent. In all experiments, we set the learning rate and the rank number to $1e^{-3}$ and 8, respectively. The $\alpha$ was chosen to be 32, and the dropout rate was 0.1. The number of update steps is 30 for GPT2-XL and 50 for GPT-J.

\paragraph{MEND}
MEND \cite{mend} conducts the editing by manipulating the language models' gradient. It trains a meta-network that employs a rank-1 decomposition of the model gradients and predicts a new rank-1 update to the corresponding model weights. In this work, we train two meta-networks using corresponding training split in the ZsRE and COUNTERFACT datasets for GPT2-XL following the default hyperparameter settings. Due to the large required computation resource for training GPT-J (6B) meta-network, we do not perform training for GPT-J. 

\paragraph{SERAC}
\citet{serac} designed a memory-augmented editing method, which requires an external cache to store explicit editing cases. It also adopts a scope classifier that determines whether an input sample falls in the editing scope and a small counterfactual model for editing the in-scope cases. For both GPT2-XL and GPT-J, we train two separate models for the two datasets, respectively. Following the original paper, we apply distilbert-base-cased \cite{distil-bert} for the scope classifier across all models. For the small counterfactual model, we employ GPT2 for GPT2-XL and gpt-j-tiny-random\footnote{https://huggingface.co/anton-l/gpt-j-tiny-random} for GPT-J. All hyperparameters follow default settings.

\paragraph{MEMIT}
MEMIT \cite{memit} treats the feedforward layer of transform as a linear associative memory and uses a minimum square error optimization to add new key-value associations to layer weights. We follow the original paper to edit the layers in the identified critical path and set the balance factor $\lambda$ to the optimal value found in the original work. Other parameters for the two models are all set based on configurations in \cite{memit, rome}.

\paragraph{GRACE}
\citet{grace} proposed an editing method that preserves the original model parameters and adopts a codebook maintained by adding, splitting, and expanding keys over time to store relevant edits. We follow the optimized settings in the original paper and set the value optimizing learning rate to 1. The number of iterations for optimizing the values is 100, and the initial $\varepsilon$ value is chosen to be 1. The codebook is employed at layers 35 and 25, respectively.

\paragraph{CoachHooK}
CoachHooK expands the update mechanism in MEMIT to consecutive cases and applies hook layers to separate the weight change from the original model layer. For both models, we set $\lambda=15,000$, $\alpha_z=2.2$ for consecutive batch editing. Unless specified, we evaluate our method on full critical path layers identified in \cite{memit}. We employ the same procedure in MEMIT \cite{memit} to compute the updating keys and the target values, except that the most recently updated model during the process of consecutive editing is applied for relevant computations.  
We applied "torch.float16" for the GPT-J model for all experiments.

\section{More detailed analysis and discussions}
\label{appendix_C}

\paragraph{Effect of the Number of Editing Layers}
\begin{figure}
\centering
\includegraphics[width=0.48\textwidth]{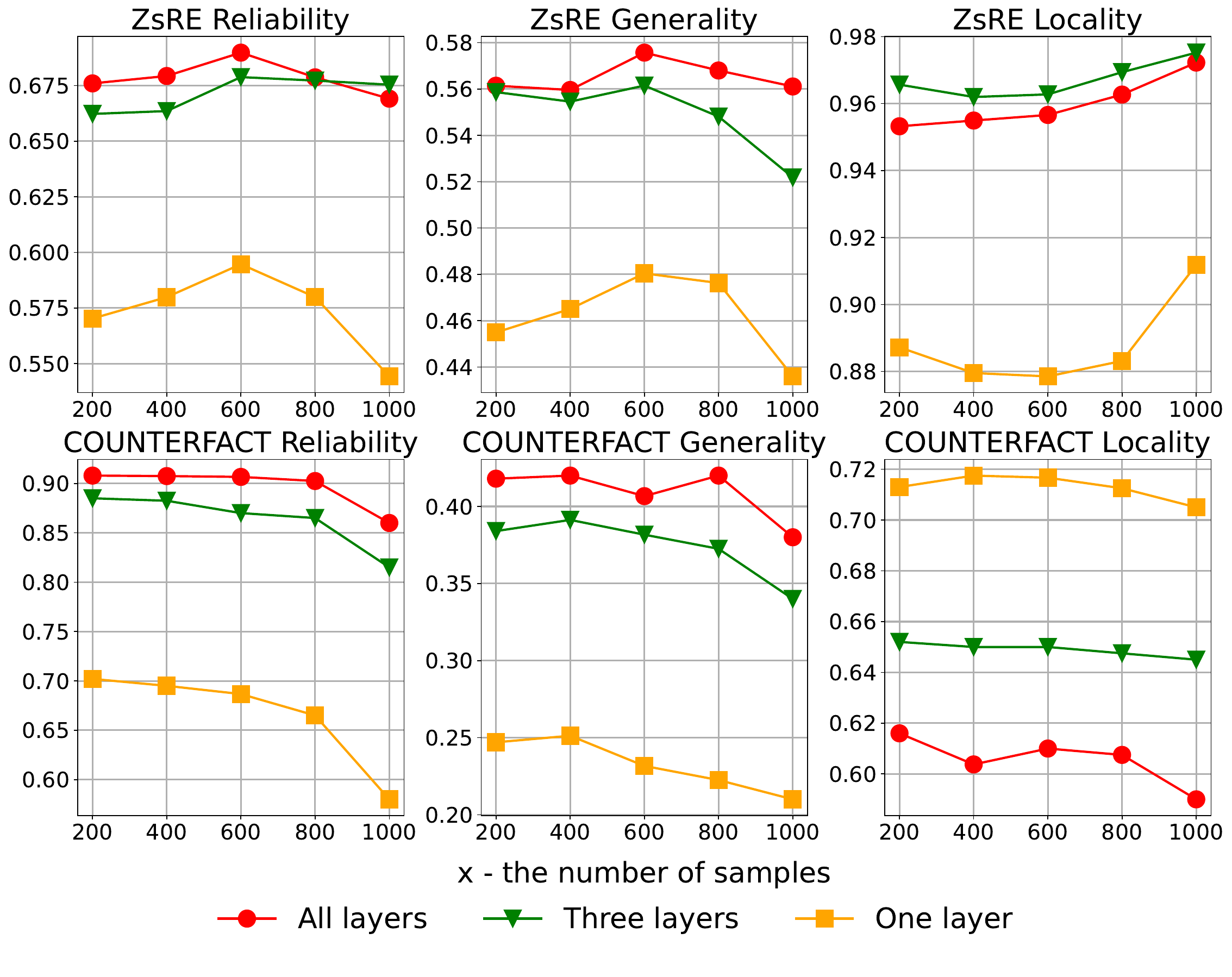}
\caption{Performance comparisons on the different number of editing layers. Layers are selected from the critical path identified in \cite{memit}.}
\label{fig:Layers_variation}
\end{figure}

To investigate the necessity of applying the hook layer onto multiple transformer layers, we conduct the consecutive batch editing experiment on the ZsRE dataset for GPT2-XL (Fig. \ref{fig:Layers_variation}). As the effect of choosing different layers has already been studied in \cite{memit}, we focus only on the effect of the number of layers. We selected the last one, three, and all layers from the critical path identified in \cite{memit, Yao}, respectively.

As shown in Fig. \ref{fig:Layers_variation}, the one-layer case significantly underperforms the other two cases in most of the metrics for the two datasets, which directly certifies the necessity of the expansion. In ZsRE, the difference between the performance for one layer and multiple layers tends to enlarge in reliability and generality as the consecutive editing steps increase. This may serve as evidence of our assumption in section \ref{sec 3.3}, which mentions that the latter hook layer may capture the in-scope instances missed in former hook layers. Additionally, the all-layer case has slightly better generality than the three-layer case, and they do not show a remarkable difference in locality and reliability. A similar situation could be found in the reliability and generality of the COUNTERFACT with an interesting exception in the locality, where an adverse performance order of the cases is shown. Nevertheless, the margin of the locality fall is not that manifest in contrast with the advancement in reliability and generality.

\paragraph{Effect of the Initial Threshold \(\alpha_z\)}
\begin{figure}[t]
\centering
\includegraphics[width=0.48\textwidth]{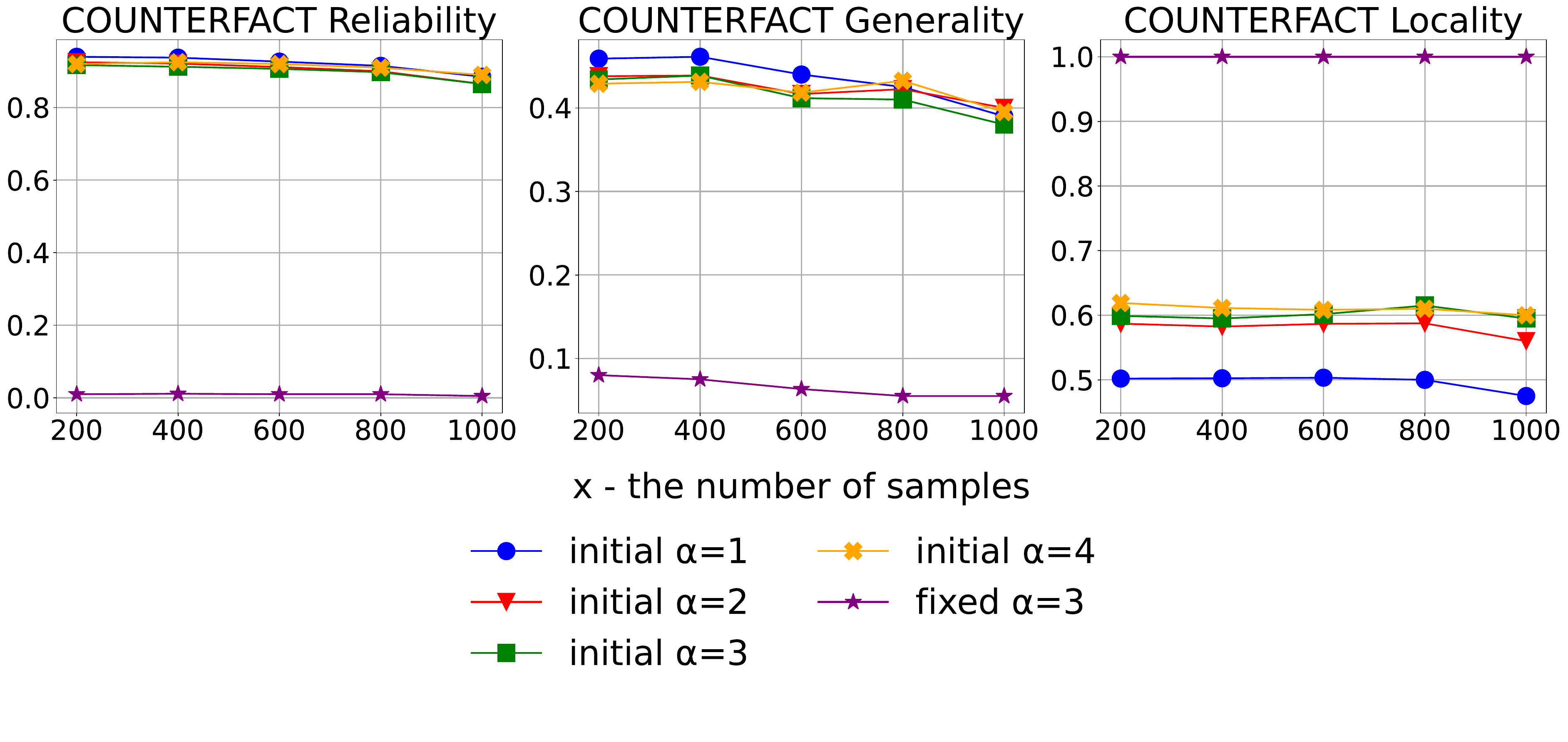}
\caption{Performance comparisons on different $\alpha_z$.}
\label{fig:vary_alpha}
\end{figure}

$\alpha_z$ is the initialization value of $\alpha$ used in the identification of local editing scope (section \ref{sec 2.3}). We study its influence in this part. According to Fig.\ref{fig:vary_alpha}, although the $\alpha_z=1$ case ranks the highest in the first 60 editing steps in generality, it consistently performs the worst in locality, indicating that it fails to intercept many out-scope inputs. This implies that 1 may be too low for the initialization. Other cases do not show noticeable differences in the three metrics since $\alpha_z$ is just the initial value and $\alpha$ is determined dynamically. It seems that overly low $\alpha_z$ would damage the hook layer's capacity to discriminate in-scope and out-scope samples. Considering the unpredictable consecutive steps that our method may be applied, we select a relatively low value between 2 and 3, namely, $\alpha_z=2.2$.

To verify the significance of the dynamical determination process, we also test the fix $\alpha$ case. We chose the value of 3, the standard threshold used in standardization to detect outliers. The results reveal a dramatic decline in reliability and generality and perfect fulfillment in the locality, indicating that almost all instances are indiscriminately obstructed by the hook layers regardless of the editing scope. Besides, choosing an optimal fixed $\alpha$ before editing is practically unrealistic. Therefore, it would be more reasonable to decide $\alpha$ dynamically.

\begin{table}[t]
    \centering
    \setlength{\tabcolsep}{1mm}{
    \begin{adjustbox}{max width=\textwidth}
    \begin{tabular}{lcccccccc}
    \toprule
     Model & Type & Inference Time (s) \\ 
     \midrule
     \multirow{2}{*}{GPT2-XL} & Pre-edit & 0.1187 \\
     & Post-edit & 0.1297 \\
     \midrule
     \multirow{2}{*}{GPT-J} & Pre-edit & 0.0762 \\
     & Post-edit & 0.0863 \\
    \bottomrule
    \end{tabular}
    \end{adjustbox}}
    \caption{Inference time analysis.}
    \label{tab:time_analysis}
\end{table}

\begin{table}[t]
\small
    \centering
    \setlength{\tabcolsep}{0.5mm}{
    \begin{adjustbox}{max width=\textwidth}
    \begin{tabular}{lcccc}
    \toprule
    \multirow{2}{*}{Model} &  \multirow{2}{*}{Granularity} & \multicolumn{3}{c}{Instances} \\ 
    \cmidrule(lr){3-5}
    
     &&  Reliability & Generality & Locality \\
     \midrule
     \multirow{3}{*}{GPT-J} & Instances & 99.00 & 97.50 & - \\
     & Overall tokens & 9.69 & 12.51 & 11.19 \\
     & Unwanted tokens & 0.38 & 0.13 & 11.19 \\
    \bottomrule
    \end{tabular}
    \end{adjustbox}}
    \caption{Percentage of instances/tokens that used the hook layer. }
    \label{tab:hook layer employment}
\end{table}

\begin{table}[t]
    \centering
    \setlength{\tabcolsep}{0.5mm}{
    \begin{adjustbox}{max width=0.48\textwidth}
    \begin{tabular}{lccccc}
    \toprule
    \multirow{2}{*}{\makecell{Deferral\\Radius}} & \multirow{2}{*}{Model} & \multicolumn{4}{c}{COUNTERFACT} \\ 
    \cmidrule(lr){3-6}
    
     & & Reliability & Generality & Locality & Average \\
     \midrule
     $\varepsilon = 1$ & \multirow{6}{*}{GPT2-XL} & 100 & 0.40 & 100 & 66.80 \\
     $\varepsilon = 3$ && 100 & 0.42 & 100 & 66.81\\
     $\varepsilon = 5$ && 100 & 0.65 & 99.50 & 66.72 \\
     $\varepsilon = 10$ && 100 & 1.80 & 93.70 & 65.17 \\
     $\varepsilon = 20$ && 100 & 18.30 & 56.60 & 58.30 \\
     $\varepsilon = 30$ && 100 & 83.90 & 7.40 & 63.77 \\
     \midrule
     $\varepsilon = 1$ & \multirow{6}{*}{GPT-J} & 100 & 0.50 & 100 & 66.83 \\
     $\varepsilon = 3$ && 100 & 0.54 & 100 & 66.85 \\
     $\varepsilon = 5$ && 100 & 0.57 & 100 & 66.86 \\
     $\varepsilon = 10$ && 100 & 0.68 & 99.60 & 66.76 \\
     $\varepsilon = 20$ && 100 & 5.00 & 93.20 & 66.07 \\
     $\varepsilon = 30$ && 100 & 31.30 & 58.90 & 63.40 \\
    \bottomrule
    \end{tabular}
    \end{adjustbox}}
    \caption{Results of GRACE with increased $\varepsilon$.}
    \label{tab:radius}
\end{table}

\paragraph{Investigation on hook layer employment}
Although the validation of the hook layer has been proved in section \ref{sec 3.4}, we conducted extra experiments to survey how many entries that should apply the hook layer indeed use the hook layer and vice versa. We investigated three granularity: instances, overall tokens, and unwanted tokens (Table \ref{tab:hook layer employment}). Suppose the number of instances is $A$, the total number of tokens for the set of instances is $T$, and there are $T'$ tokens that used the hook layer and $A'$ instances have their updated keys\footnote{Each instance only has one updated key.}  (the last subject token) use the hook layer. The instance granularity was measured by $\frac{A'}{A}$, the overall tokens granularity was calculated by $\frac{T'}{T}$, and the unwanted tokens $\frac{T'-A'}{T}$.

The results show that almost all reliability and generality instances apply the hook layer, and few unwanted tokens mistakenly use the hook layer. This again demonstrates the effectiveness of our method's editing scope identification.

\paragraph{GRACE with greater deferral radius}
Although we followed the settings found in the original paper of GRACE \cite{grace}, one may argue that the terrible generality performance of GRACE in Table \ref{tab:consecutive batch editing} is caused by the over small deferral radius ($\varepsilon$) and increasing it may help the model reach a better balance between generality and locality, then resulting in an improved overall average. Therefore, we further conducted the consecutive batch editing experiments for GRACE with several increased $\varepsilon$ on the COUNTERFACT dataset, the result is shown in Table \ref{tab:radius}.

It is not hard to find from the results that, though the results indeed show the trade-off between generality and locality, the average does not show great improvement. This proves that merely increasing the deferral radius for GRACE does not necessarily improve its overall average performance.

\paragraph{Inference Time Analysis}
As our method will introduce new hook layers to the model, we conduct an experiment to investigate its influence on the model inference. We run GPT2-XL on NVIDIA Titan GPU and GPT-J on NVIDIA A6000. Table \ref{tab:time_analysis} shows the running result for the corresponding pre-edit and post-edit models. The hook layers' employment does not seem to delay the model inference too much. This may result from the fact that the hook layers are only introduced for the small proportion of layers in the critical path, and the computation implemented in the hook layers is relatively simple.

\paragraph{Memory Analysis}
Unlike GRACE \cite{grace}, whose memory requirement grows over time and SERAC \cite{serac}, which needs extra memory for counterfactual model and scope classifier, the memory requirement of our method remains unchanged over time. Therefore, the final memory requirement is fixed no matter how many edits you make to the model. The initial memory requirement is acceptable since it is at maximum the copy of the 6 to 7 FFN projection layer weights in the model.
Specifically, the hook layers are only applied to a set of identified layers, which usually accounts for a small proportion of the whole layers. For example, the number of identified layers for GPT-J-6B is 6, which is [3, 4, 5, 6, 7, 8], and 5 for GPT2-XL, which is [13, 14, 15, 16, 17]. Furthermore, it is not compulsory to hang hook layers to all the identified layers, user can decide how many layers they want to edit. For convenience, we assume to use all the identified layers here. Take the GPTJ-6B as an example, a projection FFN layer weight dimension is $16384 \times 4096$, assuming the data type is float32, then GPU memory required by its hook layer (just a copy of itself) is approximately  $\frac{16384 \times 4096 \times 4}{1024^3} = 0.25$GB (ignore the bias). Now, the editing layers used in our approach for GPT-J-6B is [3,4,5,6,7,8], so the required memory is around $0.25\text{GB} \times 6=1.5\text{GB}$.

\end{document}